# Rethinking Backbone Design for Lightweight 3D Object Detection in LiDAR

Adwait Chandorkar    Hasan Tercan    Prof. Tobias Meisen

Institute for TMDT, University of Wuppertal, Germany

{chandorkar, tercan, meisen}@uni-wuppertal.de

## Abstract

*Recent advancements in LiDAR-based 3D object detection have significantly accelerated progress toward the realization of fully autonomous driving in real-world environments. Despite achieving high detection performance, most of the approaches still rely on a VGG-based or ResNet-based backbone for feature exploration, which increases the model complexity. Lightweight backbone design is well-explored for 2D object detection, but research on 3D object detection still remains limited. In this work, we introduce* **Dense Backbone**, *a lightweight backbone that combines the benefits of high processing speed, lightweight architecture, and robust detection accuracy. We adapt multiple SoTA 3d object detectors, such as PillarNet, with our backbone and show that with our backbone, these models retain most of their detection capability at a significantly reduced computational cost. To our knowledge, this is the first dense layer-based backbone tailored specifically for 3D object detection from point cloud data. DensePillarNet, our adaptation of PillarNet, achieves a 29% reduction in model parameters and a 28% reduction in latency with just a 2% drop in detection accuracy on the nuScenes test set. Furthermore, Dense Backbone's plug-and-play design allows straightforward integration into existing architectures, requiring no modifications to other network components.*

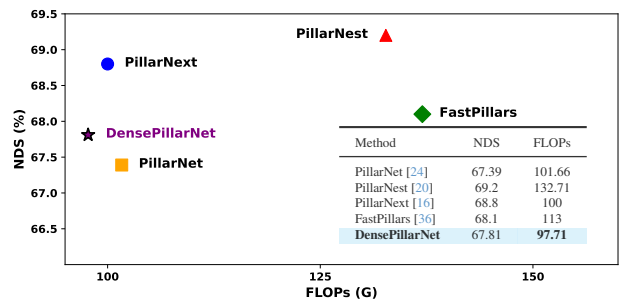

Figure 1. Comparision of SoTA Pillar-based 3d object detectors evaluated on nuScenes Detection Score (NDS) vs. GFLOPs on nuScenes *val* set. Our *Dense Backbone* adapted on PillarNet, DensePillarNet, has the lowest computational cost. Also, in comparison with other pillar-based SoTA models, there is not a substantial drop in detection accuracy.

## 1. Introduction

One of the key challenges in the autonomous driving domain is the limited availability of computational resources in the vehicle. A real-world autonomous driving vehicle must execute numerous tasks such as object detection, path planning, maneuvering and more simultaneously, in real-time and locally on an edge device. These tasks require significant computational power [2, 22], and deploying large servers within the car is impractical due to space and energy constraints. An alternative approach involves offloading computations to cloud servers; however, this solution is constrained by network reliability and bandwidth limitations, while also introducing critical challenges related to safety, data privacy, and real-time responsiveness, which are essential for autonomous driving applications. To address these constraints, machine learning models — particularly those used for object detection — must be designed to operate in real time with minimized computational requirements, while maintaining high performance.

Recent state-of-the-art (SoTA) object detection models for autonomous driving rely on 3D object detection using LiDAR point clouds. These models typically convert point clouds into voxels or pillars and employ either 3D CNNs [5, 9, 30, 35] or 2D CNNs [14, 16, 21, 24, 29, 31] for feature extraction. Models using 3D CNNs achieve state-of-the-art results on object detection task but inevitably suffer from a high computational cost which makes it challenging to deploy these models on resource-limited hardware. On the contrary, despite achieving a higher FPS and low computational cost, models using 2D CNN-based feature extractors have a slightly lower detection accuracy. Over the years, the broader research in 3D object detection has focused on improving point cloud encoding [16, 24, 36], aggregation of features in the neck [16, 24, 35] or improving the detection head to achieve better accuracy [5, 33]. However, there is very limited research in backbone design. In contrast, in 2D object detection, there has been explicit research on designing lightweight models [6, 8, 12, 13, 23, 27, 28], which are both fast and demonstrate improved detection accuracy. The feature extractors in these methods, i.e. the *backbone*, were specially designed using for example depthwise-separable convolutions [6, 12, 23] or dense layers [27, 28]. For 3D object detectors however, this area of research remains largely unexplored. The SoTA 3D object detectors that use 2D CNN-based backbone do not design their backbones explicitly, but instead use some popular image-based networks such as ResNet [11]. We posit that the drop in detection accuracy observed in these models largely stems from the reliance on image-based object detectors that lack adaptation for point cloud data, which are inherently sparse. Therefore, it becomes imperative to revisit backbone design strategies that are specifically tailored to effectively learn sparse point cloud representations, while maintaining a low computational footprint.

In this paper, we introduce a novel dense layer-based backbone network for 3D object detection, named **Dense Backbone**. Drawing inspiration from DenseNet [13], PeleeNet [28] and VovNet [15], we argue that efficient feature map reuse is crucial when learning representations from point clouds due to their inherent sparsity and lack of structure. The core component of our backbone are dense layers with one-shot aggregation that provides multiple receptive fields, thereby enhancing the expressiveness of the learned features.

A key advantage of the proposed Dense Backbone is its plug-and-play capability. We define plug-and-play as the ability to seamlessly integrate our backbone within the existing architecture without the need to modify other components such as encoder, neck, and head. To demonstrate this, we implemented our backbone on the PointPillars [14], CenterPoint [33] and PillarNet [24] frameworks [14]. We refer the adapted versions as DensePointPillars, DenseCenterPoint and DensePillarNet. DensePointPillars are evaluated on the KITTI dataset while DenseCenterPoint and DensePillarNet are evaluated on nuScenes dataset. When evaluated on the nuScenes dataset, our DensePillarNet has the lowest computational demand among recent SoTA models (using ResNet-based backbone) without a substantial drop in performance, as shown in Figure 1. Our contributions are as follows:

- We present **Dense Backbone**, a lightweight backbone that is designed to maximize feature reuse, achieving competitive detection accuracy with reduced computational costs.
- Our Dense Backbone is plug-and-play capable, meaning it is possible to replace the backbone of an existing network with Dense Backbone without modifying any other component.

To the best of our knowledge, we are the first to propose a fully dense layer-based backbone specifically for LiDAR-based 3D object detection.

## 2. Related Work

### 2.1. LiDAR-based 3D object detection

Due to the sparsity of point clouds, the 3D object detectors initially transform them into a structured representation before performing feature extraction. One basic approach is bird's-eye-view (BEV) transformation, as proposed by Chen et al. in MV3D [3]. A different methodology was proposed by Simon et al. in their work, Complex YOLO [25]. It leverages a complex region proposal network (RPN) following a BEV transformation to enhance model speed and accuracy; however, these approaches suffer significant information loss during BEV transformation. A more widely adopted method is grid-based transformation, dividing point clouds into either 3D voxels or 2D pillars. For feature extraction, Voxel-based methods [5, 9, 31, 35, 37] partition point clouds into 3D voxels using either 3D CNNs as proposed by Zhou et al. in VoxelNet [37] or 3D SpConv proposed by Yan et al. in SECOND [31]. A recent voxel-to-object method VoxelNext [5], introduced by Chen et al., proposes direct end-to-end 3D object detection from voxel features, bypassing the need for post-processing. Other approaches, such as [4, 30], utilize RGB image features to improve point cloud feature representation. A recent work SAFDNet [35], by Zhang et al., introduces sparse adaptive feature diffusion to optimize both detection accuracy and inference time.

Despite advancements in reducing computational demands, voxel-based detectors still entail high computational costs and extended inference times. On the other hand, pillar-based approaches, such as Pointpillars [14] or HVPR [21] eliminate the necessity for 3D convolutions during feature extraction. Avoiding the voxel-based methods and instead projecting point clouds into a pseudo-image format, PointPillars achieved a favorable balance between inference speed and detection accuracy, particularly on the KITTI benchmark. PillarNet [24], by Shi et al., introduced a specialized pillar encoder optimized for pseudo-image representations, coupled with a high-capacity feature aggregation neck that enhances multi-scale feature fusion. It not only outperforms [14] in accuracy but also nearly doubles processing speed. A recent work PillarNext [16], by Li et al., employs enlarged receptive fields inspired by 2D object detection, achieving superior performance over both pillar- and voxel-based models. While recent methods have enhanced pillar-based object detectors, they continue to rely on ResNet-based backbones with only minor architectural modifications. This underscores the need for further research into designing specialized backbones tailored specifically for point cloud-based 3D object detection.

## 2.2. Backbone Network

A feature extraction network, commonly referred to as *backbone* in 3D object detection, is responsible for extracting multi-level features from point clouds. These features are subsequently fused and refined by the *neck*, while the *head* detects the 3D objects. Although 3D CNNs were initially popular, they were replaced due to their high computational demands and latency. BEV-based methods [25, 26, 32] and early pillar-based models [14, 21] rely on 2D CNN backbones, often using ResNet-based backbones.

However, these backbones were not specifically designed for point clouds, nor did the models adapt their architectures to handle the sparsity and unstructured nature of point clouds. Consequently, recent voxel and pillar-based models have predominantly adopted SpConv-based backbones. Current approaches, such as [33–35], utilize submanifold SpConv Residual Blocks (SRB) followed by sparse encoder-decoder layers to capture long-range dependencies efficiently. Some pillar-based models [16, 24] employ a modified ResNet-18 [11] as a backbone, replacing conventional 2D CNNs with 2D SpConvs. Voxel-based models like [5] use a fully SpConv architecture, eliminating the need for post-processing steps, such as anchors or RPNs. While SpConv-based backbones have proven effective, recent research has shown increased interest in alternative backbone designs. FastPillars [36] by Zhou et al. introduces a novel pillar encoding technique, showing that, when combined with a tailored 2D CNN-based backbone, it can achieve enhanced detection accuracy with reduced computational load and improved speed. PillarNest [20] by Mao et al. proposes a ConvNext [17] inspired backbone which is first pre-trained on large-scale image datasets such as ImageNet and then scaled to adapt the features of point cloud. Both approaches however, require some modification in the pillar encoder as well as in the loss function [20] to incorporate the new backbone. We demonstrate that our backbone is fully plug-and-play and can be easily adapted in the network without modifying any other component.

## 2.3. Lightweight network design

With advancements in 2D object detectors, implementing models on resource-constrained hardware emerged as a critical challenge. Beyond traditional techniques like *pruning* and *quantization*, several researchers have introduced an approach known as *Lightweight network design* to reduce model size while retaining performance. Chollet introduced Xception [6] employing depthwise separable convolutions. Inspired by Xception's success, Howard et al. presented MobileNet [12] and Sandler et al. expanded it further with MobileNetv2 [23], both utilizing depthwise separable convolutions for object detection. These models demonstrated that lightweight, efficient detectors could meet the computational requirements of mobile and embedded applications.

Alternatively, Huang et al. propose DenseNet [13], using conventional 2D CNNs, and emphasize feature reuse to improve the model performance while simultaneously reducing the model parameters. DenseNet creates dense connectivity by concatenating feature maps from all preceding layers into each new layer, rather than summing them as in ResNet [11]. This structural difference allows DenseNet to achieve richer feature reuse across the network, contributing to improved efficiency.

While DenseNet initially targeted image classification, object detection presents additional challenges, including handling multiple objects with varying classes and aspect ratios. Addressing these complexities, Wang et al. developed PeleeNet [28] for 2D object detection, which modified DenseNet by introducing a two-way dense layer to expand receptive fields, enhancing the model's spatial context awareness. This design allows PeleeNet to outperform other lightweight object detectors, such as MobileNet, while maintaining lower FLOPs and latency. However, due to the concatenation of features in each layer, both PeleeNet and DenseNet suffer from heavy memory access costs. VovNet [15] by Lee et al. proposed an energy and deployment efficient network by replacing layerwise concatenation with one shot aggregation after multiple layers.

By incorporating *dense layers* and *one-shot aggregation* into *Dense Backbone*, we achieve the spatial depth and efficiency needed for efficient 3D object detection at a marginal computational cost, particularly in scenarios constrained by computational resources.

## 3. Dense Backbone

### 3.1. Design Rationale

Our objective is to minimize model parameters and computational cost by strategically reducing inter-layer connections. This can typically be achieved by increasing the stride or reducing the output channels in convolutional layers. However, higher strides tend to decrease spatial resolution and limit receptive field overlap, thereby impeding the capture of fine-grained features. Similarly, reducing the number of channels may hinder the model's ability to represent details if each channel fails to capture essential feature distinctions. This motivated our approach: first, to reduce the number of output channels to lower computational cost; and second, to mitigate the resulting loss in representational capacity by maximizing feature reuse across layers. This enables the model to effectively capture features at multiple spatial scales, effectively rethinking the conventional design of the backbone.

### 3.2. Design

**Dense Block** - The building block of Dense Backbone is the dense layer. Inspired by VoVNet [15], we employ a series

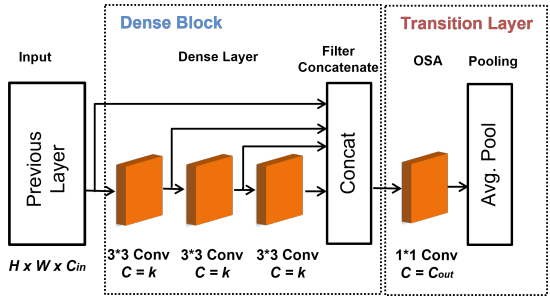

Figure 2. Dense Backbone Architecture. A *Dense Block* consists of multiple cascaded dense *3\*3* Conv layers. The transition layer aggregates the concatenated features from the *Dense Block* with a pointwise Conv layer followed by Pooling. The output channels of *3\*3* Conv are controlled by a hyperparameter *growth_rate*, denoted by $k$.

of feed-forward Conv layers followed by concatenation at the end as shown in Figure 2. The concatenation of features enables simultaneous feature learning for objects with varying aspect ratios.

**Transition Layer** - The transition layer is an intermediate layer between subsequent *dense blocks*. To avoid reducing the receptive field of the concatenated features we aggregate these features with a point-wise convolution layer. Finally, we incorporate an average pooling layer with a stride of 2. Our extensive studies prove that *pooling* layer offers better detection accuracy over strided convolutions.

**Growth Rate** - The *growth_rate* ($k$) determines the number of learnable parameters in each dense layer. Fewer channels in the 3x3 kernels certainly help in reducing network weight, but this can also impact feature learning. To overcome this, unlike DenseNet [13] which adopts a standard growth rate of $k = 32$, we modify $k$ for each *dense block* to ensure there is no large disparity in concatenated channels and output channels $C_{out}$. We initialize the growth rate $k$ to 32 in the first *Dense Block*, following the design choices in [13, 28], and progressively double it in each subsequent *Dense Block*. This strategy allocates more learnable parameters to deeper layers, enhancing their capacity to extract higher-level semantic features.

### 3.3. DensePointPillars

Figure 3 represents the overall architecture of DensePointPillars, which integrates our Dense Backbone into the original PointPillars [14] framework.

**Encoder**: Like the original PointPillars model, DensePointPillars uses a grid-based encoder that organizes raw points into pillars. These pillars are then stacked, and multi-layer perceptrons (MLPs) extract features from them. The extracted features are transformed into a 2D pseudo-image for further processing.

**Backbone**: Due to the plug-and-play nature of our Dense Backbone, we maintain the original network's backbone configuration, including the number of layers and output channels to the *neck*. Each Dense Block in the backbone consists of a specific number of dense layers, arranged as [3,5,5] similar to the structure in the original paper, with three layers in the first block and five layers in each of the subsequent blocks.

**Neck**: The *neck* module aggregates the features from the backbone, fusing outputs from the *transition layers* to incorporate multi-scale context and expand the receptive field. This approach enhances the detection head's ability to accurately localize and classify objects of various sizes. We employ a Feature Pyramid Network (FPN) as *neck* as proposed in the original paper.

**Detection Head**: The detection head is responsible for localizing bounding boxes around objects and classifying them. As proposed in SECOND [31] and PointPillars [14], we use an anchor-based detection head with predefined anchor shapes specifically designed for 3D object detection. We retain the default detection head settings as defined in PointPillars.

### 3.4. DensePillarNet

We adapt our Dense Backbone in the PillarNet-18 [24] model as well. The original PillarNet-18 used ResNet-18 as the backbone, but replaced the Conv2d with SpConv. We follow a similar approach, retaining the *Dense Block* and *Transition Layer* layout while replacing the Conv2d in Figure 2 with SpConv. Additionally, we adopt strided convolutions in place of average pooling, as proposed in original PillarNet. This design choice is motivated by implementation constraints—specifically, sparse convolution libraries such as SpConv do not natively support pooling operations, and incorporating them would require custom extensions. The other components such as *Encoder*, *Neck* and the *Detection Head* are the same as the original PillarNet-18.

## 4. Experiments

### 4.1. Selecting the Baseline

To rigorously evaluate the plug-and-play capability and generalizability of our proposed *Dense Backbone*, we integrate it into three representative and widely adopted 3D object detection frameworks. First, we select PointPillars [14] as the *base* for experiments on the KITTI dataset given its simplicity, wide adoption, and strong performance. Second, we adopt CenterPoint [33] as the *base* model for the nuScenes dataset. CenterPoint's detection head, *CenterHead*, has become the de facto standard for recent LiDAR-based detection pipelines [5, 16, 24, 36] due to its versatility and superior performance. Leveraging CenterPoint as our baseline therefore allows us to fairly assess our backbone's adaptability and effectiveness on this widely adopted de-

1701

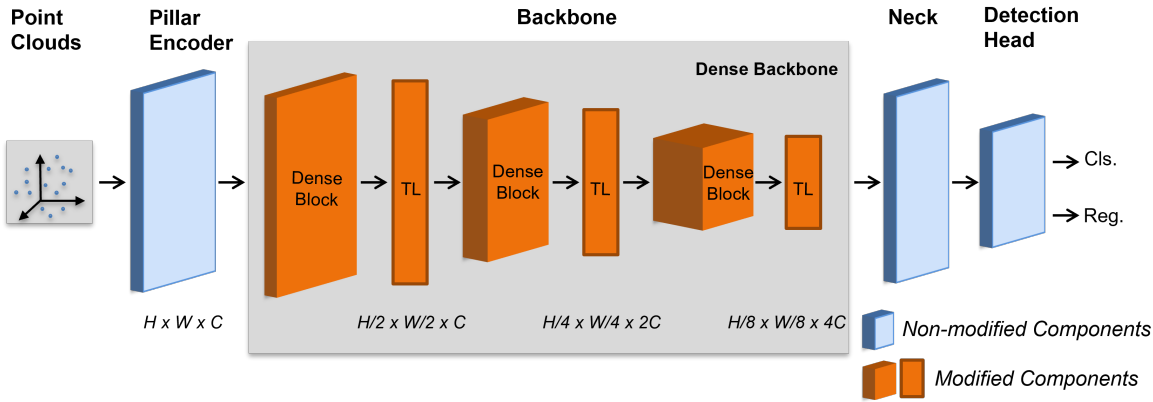

Figure 3. Network Architecture of DensePointPillars. All the components other than the backbone are same as mentioned in the Pointpillars paper [14]. Our Dense Backbone consists of three Dense Blocks and a Transition layer between two dense blocks. The features extracted in the Dense blocks are aggregated in the Transition layer using 1*1 convolution and an average pooling layer. The output of each transition layer is forwarded to the Neck.

tection method. Finally, we adapt our Dense Backbone into PillarNet [24]. PillarNet is a strong baseline for pillar-based detectors and is also used as a benchmark in recent works [16, 20, 36].

### 4.2. Datasets

We evaluate our approach on the KITTI 3D and nuScenes datasets. The KITTI 3D object detection dataset, based on Velodyne 64 LiDAR, captures scenes at a frequency of 10 Hz. We replicate the train-validation split proposed in [4, 14, 30], using 3712 samples for training and 3769 for validation. For test server submissions, we adhere to the same setup, creating a mini-validation set with 741 samples and using the remaining data for training. Performance is measured by the 3D Average Precision (AP) metric across 40 recall thresholds (R40), in line with the official KITTI benchmark. The Intersection-over-Union (IoU) thresholds are set to 0.5 for the *Pedestrian* and *Cyclist* classes, and 0.7 for the *Cars* class.

In the nuScenes evaluation, we utilize the full dataset, which consists of 1000 scenes, each approximately 20 seconds long, captured by a 32-beam LiDAR sensor operating at 20 Hz, resulting in 390k LiDAR sweeps. Following the mmdetection3d [7] configuration, we use 28,130 samples for training and 6019 for validation. Our evaluation metrics include the mean Average Precision (mAP) across ten classes, with distance matching thresholds set at 0.5, 1, 2, 4 meters. We also report the nuScenes Detection Score (NDS), which combines mAP with additional metrics assessing translation error (mATE), scale error (mASE), orientation error (mAOE), velocity error (mAVE), and classification accuracy (mAAE), providing a comprehensive measure of overall detection performance.

### 4.3. Setup details

All models were trained from scratch in PyTorch [1] using the mmdetection3d framework [7]. For the **KITTI** dataset, following the specifications in [10], we set the detection range to [0, 69.12] along the x-axis, [-39.68, 39.68] along the y-axis, and [0, 1] along the z-axis. The pillar size and the maximum number of points per pillar in PointPillars [14]. Unlike the original PointPillars model, which trained separate models for *Car* and *Pedestrian-Cyclist* detection, the mmdetection3d trains a single model across all three classes. We adopt this unified model setup in mmdetection3d [7] as our baseline. DensePointPillars, similarly, is trained on all three classes simultaneously. Consequently, the detection accuracy of our results may vary slightly from those reported in the original paper.

For the **nuScenes** dataset, to demonstrate the plug-and-play adaptability of our backbone, we implement a CenterPoint [33] as well as a PillarNet [24] adaptation, with our Dense Backbone. For DenseCenterPoint, we set the detection range to [-51.2, 51.2] horizontally and [-5, 3] vertically, with a maximum of 20 points per pillar. For detection, we use the CenterHead configuration as specified in [33]. For DensePillarNet, just like the original setting, we set the detection range to [-54, 54] horizontally and [-5, 3] vertically. We set the pillar size to (0.075, 0.075).

**Training**: On KITTI, we train our models for 80 epochs on four Tesla A100 GPUs with a batch size of eight. We used *AdamW* [19] optimizer and an initial learning rate of 0.001, employing *Cosine Annealing* [18] for scheduling. During validation, we set the NMS threshold to 0.01 to filter out irrelevant bounding boxes. For nuScenes, we train our models for 20 epochs on four Tesla A100 GPUs with a batch size of two, using the same optimizer and scheduler setting as for KITTI but the initial learning rate of 0.0001. The NMS threshold is set to 0.2 for all the classes, as in [5, 16, 35]. For DensePillarNet, however, we keep the NMS threshold at 0.1 and use only double-flip ensembling as stated in the original paper [24].

| Method | mAP↑ | Car 3D AP (R40)↑ | | | Ped. 3D AP (R40)↑ | | | Cyc. 3D AP (R40)↑ | | | Time↓ |
|---|---|---|---|---|---|---|---|---|---|---|---|
| | | Easy | Mod. | Hard | Easy | Mod. | Hard | Easy | Mod. | Hard | (ms) |
| 3D object detection | | | | | | | | | | | |
| PointPillars [14]* | 55.44 | 83.51 | 73.13 | 68.02 | 39.38 | 32.10 | 29.54 | 70.51 | 54.47 | 48.34 | **16** |
| DensePointPillars | **57.62** | **84.60** | **75.46** | **68.43** | **42.76** | **35.38** | **32.63** | **70.95** | **57.36** | **50.98** | 19 |
| BEV object detection | | | | | | | | | | | |
| PointPillars [14]* | 63.31 | 91.25 | 85.96 | 80.82 | 45.74 | 38.07 | 35.76 | **75.76** | 61.70 | 54.77 | **16** |
| DensePointPillars | **64.47** | **92.13** | **86.31** | **81.12** | **47.80** | **40.31** | **37.49** | 75.14 | **63.27** | **56.70** | 19 |

Table 1. Comparison of DensePointPillars vs Pointpillars [14] on KITTI *test* set for Car, Pedestrian and Cyclist class. AP is calculated for 40 Recall values at different difficulty levels. Unlike experiments in the paper, the *base* model as well as our model is trained end-to-end on all three classes at once. * Our baseline implemented in [7].

| Method | Car | Truck | Bus | Trail. | CV | Ped. | MC | Bic. | TC | Bar. | mAP | NDS |
|---|---|---|---|---|---|---|---|---|---|---|---|---|
| CenterPoint-Second [7] | **83.9** | **49.5** | **61.9** | 34.1 | **28.3** | 76.9 | 44.1 | 18.0 | 54.0 | 59.1 | 49.4 | 59.8 |
| DenseCenterPoint | 82.1 | 45.0 | 50.4 | **42.9** | 18.3 | **77.9** | **48.9** | **18.5** | **64.8** | **63.6** | **51.2** | **60** |
| PillarNet-18 [24] | **87.4** | **56.7** | 60.9 | **61.8** | **30.4** | **87.2** | **67.4** | **40.3** | **82.1** | **76.0** | **65.0** | **70.8** |
| DensePillarNet | 86.9 | 55.8 | **65.0** | 61.0 | 23.4 | 85.6 | 63.6 | 36.6 | 80.0 | 73.7 | 63.13 | 69.4 |

Table 2. Comparison on nuScenes *test* dataset for all 10 classes. Additionally, we show the mean Average Precision (mAP) over all classes and the benchmark metric NDS. Abbreviations: Trail.: Trailer, CV: Construction Vehicle, MC: Motorcycle, Ped.: Pedestrian, Bic.: Bicycle, TC: Traffic Cone, Bar.: Barrier.

| Method | NDS↑ | mAP↑ | mATE↓ | mASE↓ | mAOE↓ | mAVE↓ | mAAE↓ | Params (M)↓ | FLOPs (G)↓ | Latency (ms)↓ |
|---|---|---|---|---|---|---|---|---|---|---|
| PillarNet [24] | 67.39 | 59.90 | 0.28 | 0.25 | 0.29 | **0.25** | 0.19 | 14.57 | 101.66 | 117 |
| PillarNeSt [20] | **69.2** | **63.2** | 0.27 | 0.25 | 0.27 | 0.27 | 0.19 | **8.9** | 132.71 | - |
| PillarNext [16] | 68.8 | 62.5 | 0.28 | 0.25 | 0.27 | 0.25 | 0.20 | 14 | 100 | 103* |
| FastPillars [36] | 68.2 | 61.3 | 0.28 | 0.25 | 0.26 | 0.25 | 0.20 | 23.28 | 136.81 | 113 |
| DensePillarNet | 67.81 | 60.33 | **0.27** | **0.25** | **0.27** | 0.26 | **0.19** | 10.35 | **97.71** | **85** |

Table 3. Comparison of our model with other pillar-based SoTA on nuScenes *val* set. mATE, mASE, mAOE, mAVE and mAAE denote the errors associated with translation, scale, orientation, velocity and classification accuracy.
*The latency in the original paper was recorded on a faster hardware.

## 5. Results

### 5.1. KITTI dataset

Table 1 presents the results of DensePointPillars on the KITTI *test* set. Our model achieves an improvement of 1-2% in detection accuracy on all classes for 3D as well as BEV tasks compared to the *base* model. In addition to the promising results, our model requires 33% fewer computa-tions and has four times fewer parameters compared to the *base* model. Despite this, our model has slightly higher latency as compared to the *base* which can be attributed to its increased memory demand from concatenation, which necessitates a high global memory bandwidth utilization for copying input tensors into a new, contiguous output.

Nevertheless, DensePointPillars still achieves a speed of more than 52 Hz which is more than twice as fast as LiDAR's operating speed of 20 Hz.

### 5.2. nuScenes dataset

Table 2 presents a class-wise performance comparison between our model and the *base* model on the nuScenes

1703

dataset. DenseCenterPoint achieves superior performance in most classes, with a 2% improvement in mAP and a slight increase in NDS over CenterPoint [33]. Although DensePillarNet does not outperform the *base* model, its performance is more or less similar across most classes and there is just 1.5% drop in NDS.

Table 3 provides results on additional true positive metrics in the nuScenes *val* set. We compare the results not only with the baseline but also with a couple of new SoTA models such as PillarNest [20], FastPillars [36] and PillarNext [16], which are an improvement over our base model. With Dense Backbone, we observe that the tracking and orientation errors are either similar or slightly improved, contributing to the overall increase in the NDS score. But most importantly, our model needs significantly fewer computations than the latest SoTA model and thereby achieves better speed. The results demonstrate that integrating our Dense Backbone has a reduced computational demand and thereby faster inference speed, giving it an advantage for edge device deployment.

## 6. Ablation studies

### 6.1. Lightweight Model

Table 4 provides a detailed breakdown of the model parameters and computational demands for each component of DensePointPillars on the KITTI dataset and DensePillarNet on the nuScenes dataset. When adapted to PointPillars, our *Dense Backbone* has 9x fewer parameters resulting in 1.5x fewer FLOPs compared to *base*, whereas for PillarNet, our *Dense Backbone* has 2.5x fewer parameters resulting in 2.5x fewer FLOPs. In both cases, the results of our models are comparable to *base*, which suggests that despite the lesser computations, the model learns important features. These findings support our hypothesis that for sparse, unstructured data such as LiDAR, a backbone optimized for feature reuse is preferable to a larger backbone with limited reuse.

### 6.2. Impact of Growth Rate

*Growth rate* ($k$) controls the flow of information through the network, influencing both model parameters and computational demands. In their original work, Huang et al. [13] suggested a fixed growth rate of 32. On the other hand, Lee et al. [15] proposed a stage-wise adjusted $k$ as they replaced layer-wise concatenation with one-shot aggregation. As shown in Table 5, using a fixed growth rate $k$ slightly reduces the computational load but leads to a minor decline in detection accuracy. In contrast, progressively increasing $k$ across Dense Blocks incurs a modest increase in computation while yielding consistent gains in detection performance. Given that the accuracy improvements outweigh the added cost, we adopt the varying-$k$ strategy in our final design.

|  | PointPillars [14] | | DensePointPillars | |
|---|---|---|---|---|
| Component | Params | FLOPs | Params | FLOPs |
| Backbone | 4.21 | 29.71 | **0.47** | **19.86** |
| Neck | 0.6 | 3.13 | 0.6 | 3.13 |
| Head | 0.03 | 1.49 | 0.03 | 1.49 |

(a) KITTI dataset

|  | PillarNet [24] | | DensePillarNet | |
|---|---|---|---|---|
| Component | Params | FLOPs | Params | FLOPs |
| Backbone | 1.78 | 6.47 | **0.69** | **2.52** |
| Neck | 7.51 | 69.93 | 7.51 | 69.93 |
| Head | 1.75 | 25.21 | 1.75 | 25.21 |

(b) nuScenes dataset

Table 4. Component-wise decomposition of parameters and FLOPs on KITTI and nuScenes.

| Growth Rate | NDS↑ | Params↓ (M) | FLOPs↓ (G) |
|---|---|---|---|
| 16 | 65.69 | 9.44 | 95.42 |
| 32 | 66.18 | 9.64 | 95.66 |
| 64 | 67.38 | 10.24 | 96.22 |

(a) Growth Rate Value

| Growth Rate type | NDS↑ | Params↓ (M) | FLOPs↓ (G) |
|---|---|---|---|
| Fixed | 67.38 | 10.24 | 96.22 |
| Stagewise | **67.81** | 10.35 | 97.71 |

(b) Growth Rate type

Table 5. Impact of Growth Rate on nuScenes *val* set.

### 6.3. Runtime Analysis

| Model | A100 32GB | | Jetson Orin Nano 8GB | |
|---|---|---|---|---|
|  | Mem (MB) ↓ | FPS ↑ | Mem (MB) ↓ | FPS ↑ |
| PointPillars [14] | 250 | 60 | 25 | 12 |
| DensePointPillars | 600 | 51 | 70 | 9 |
| CenterPoint [33] | 520 | 7 | - | - |
| DenseCenterPoint | 400 | 7.5 | 40 | 2 |

Table 6. Runtime analysis of our models on KITTI and nuScenes datasets. We compare the results on NVIDIA A100 GPU as well as Jetson Orin Nano, which is an edge device.

Table 6 presents a comparative runtime analysis between our models and their respective *base* architectures. On the KITTI dataset, *DensePointPillars* exhibits a marginal increase in runtime compared to its *base*, primarily due to higher memory usage from dense feature concatenation. Nevertheless, it achieves 9 FPS on the Jetson Orin Nano, indicating near real-time performance. On the nuScenes dataset, where each sample is substantially larger and imposes higher memory demands, *DenseCenterPoint* demonstrates both faster inference and lower memory usage compared to the *base* model. Notably, while our model runs at 2 FPS on the edge device, the *base* model fails to execute due to memory overflow, highlighting the practicality of our approach in memory-constrained deployment scenarios.

## 7. Conclusion and Future Work

### 7.1. Conclusion

In this paper, we present *Dense Backbone*, a lightweight backbone built by leveraging dense connections, providing an alternative to the currently prevalent VGG-based or ResNet-based backbones for 3D object detection. To mitigate the high memory access costs typically associated with dense connections, we adopt an efficient one-shot aggregation strategy, enabling practical deployment without compromising runtime efficiency. Our design focuses on efficient feature reuse, capturing both fine-grained and high-level information across layers, which we demonstrate as essential for achieving robust 3D detection accuracy. When adapted on multiple SoTA models, *Dense Backbone* enables the models to achieve competitive detection accuracy while significantly reducing model size and computational overhead. Notably, *Dense Backbone* is fully plug-and-play, requiring no modifications to other components within existing detection frameworks.

**Limitations**: Although feature reuse improves the capture of fine-grained representations, a key limitation is increased memory usage. Our experiments conclude that one-shot aggregation is much more effective in reducing computational overhead compared to layerwise concatenation but still remains memory intensive. This higher memory demand may pose challenges for deployment on devices with extreme memory constraints, as seen in the results for Nvidia Jetson Orin Nano, potentially limiting its applicability in highly constrained environments.

### 7.2. Future Work

Although DenseBackbone offers several benefits, there is potential to further reduce its computational cost. Future exploration should investigate the combination of our approach with modifications in *Neck* and/or *Encoder*, as proposed in FastPillars [36], to optimize computational efficiency without sacrificing performance. An alternative approach could be to incorporate knowledge distillation techniques to further improve the detection accuracy while retaining the lightweight structure. We believe that our approach introduces a new perspective on backbone design and could serve as a foundation for advancing 3D object detectors that utilize dense layer-based backbones.

## Acknowledgements

This work has received funding from the European Union's Horizon Europe Research and Innovation Programme under Grant Agreement No 101076754 - AIthena project.